\documentclass[conference]{IEEEtran}
\IEEEoverridecommandlockouts
\usepackage{cite}
\usepackage{amsmath,amssymb,amsfonts}
\usepackage{algorithmic}
\usepackage{graphicx}
\usepackage{subcaption}
\usepackage{float}
\usepackage{textcomp}
\usepackage[table,xcdraw]{xcolor}
\usepackage{url}
\usepackage{lscape}
\usepackage{booktabs}
\usepackage{multicol}
\usepackage{tabularx}
\usepackage{afterpage}
\usepackage{placeins}

\def\BibTeX{{\rm B\kern-.05em{\sc i\kern-.025em b}\kern-.08em
    T\kern-.1667em\lower.7ex\hbox{E}\kern-.125emX}}

\begin{document}

\title{Pokémon Red via Reinforcement Learning\\
}

\author{
\IEEEauthorblockN{Marco Pleines\IEEEauthorrefmark{1}, Daniel Addis\IEEEauthorrefmark{3}, David Rubinstein\IEEEauthorrefmark{2},\\
 Frank Zimmer\IEEEauthorrefmark{5}, Mike Preuss\IEEEauthorrefmark{6}, Peter Whidden\IEEEauthorrefmark{2}}
\IEEEauthorblockA{
\IEEEauthorrefmark{1}\textit{TU Dortmund University}, Dortmund, Germany \\
\IEEEauthorrefmark{2}\textit{Independent Researcher}, Brooklyn, NY, USA,
\IEEEauthorrefmark{3}\textit{Independent Researcher}, Minneapolis, MN, USA\\
\IEEEauthorrefmark{5}\textit{Rhine-Waal University of Applied Sciences}, Kamp-Linfort, Germany \\
\IEEEauthorrefmark{6}\textit{LIACS Universiteit Leiden}, Leiden, Netherlands \\
}
}

\maketitle


\begin{abstract}
We present a Deep Reinforcement Learning (DRL) agent that successfully completes the first several hours of Pokémon Red, a classic Game Boy JRPG that exposes significant challenges as a testbed for agents, including multitasking, long horizons of tens of thousands of steps, hard exploration, and a vast array of potential policies.
Our agent completes an initial segment of the game, up to Cerulean City, a location requiring progression through two cities, battle-filled passages, a maze-like cave, and defeating the first gym leader.
Our experiments include various ablations that reveal vulnerabilities in reward shaping.
We argue that long-form games like Pokémon hold strong potential for future research, presenting long-term coherent reasoning challenges absent from simpler arcade games.
Our environment wrapper, training algorithm, human replay data, and pretrained agent are available at (REDACTED FOR REVIEW).

\end{abstract}

\section{Introduction}
\makeatletter
\newcommand*{\rom}[1]{\expandafter\@slowromancap\romannumeral #1@}
\makeatother

\begin{figure*}
    \centering
    \begin{subfigure}[b]{0.24\textwidth}
        \includegraphics[width=\textwidth]{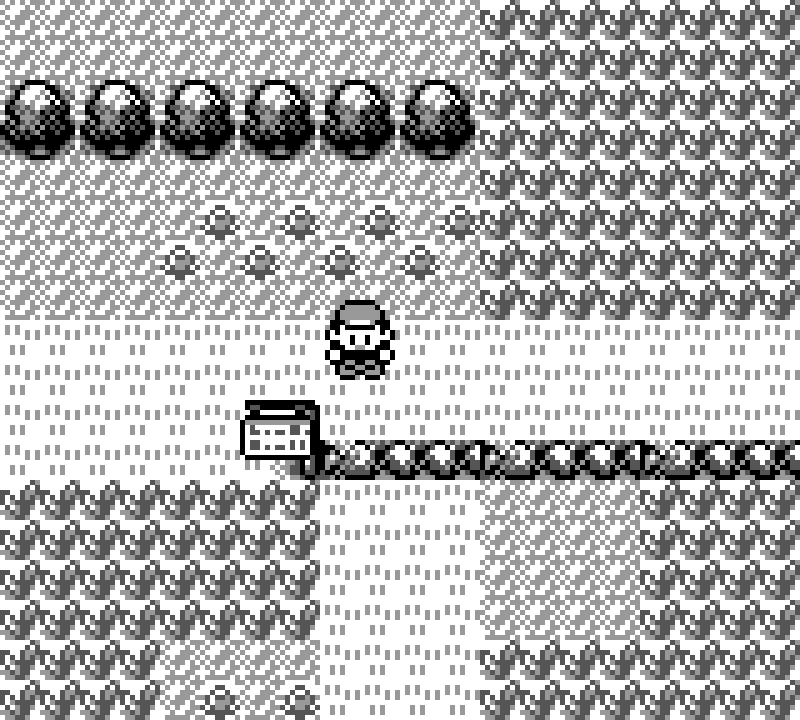}
        \caption{Overworld with Long Grass outside of Pallet Town}
        \label{fig:route1}
    \end{subfigure}%
    \hfill
    \begin{subfigure}[b]{0.24\textwidth}
        \includegraphics[width=\textwidth]{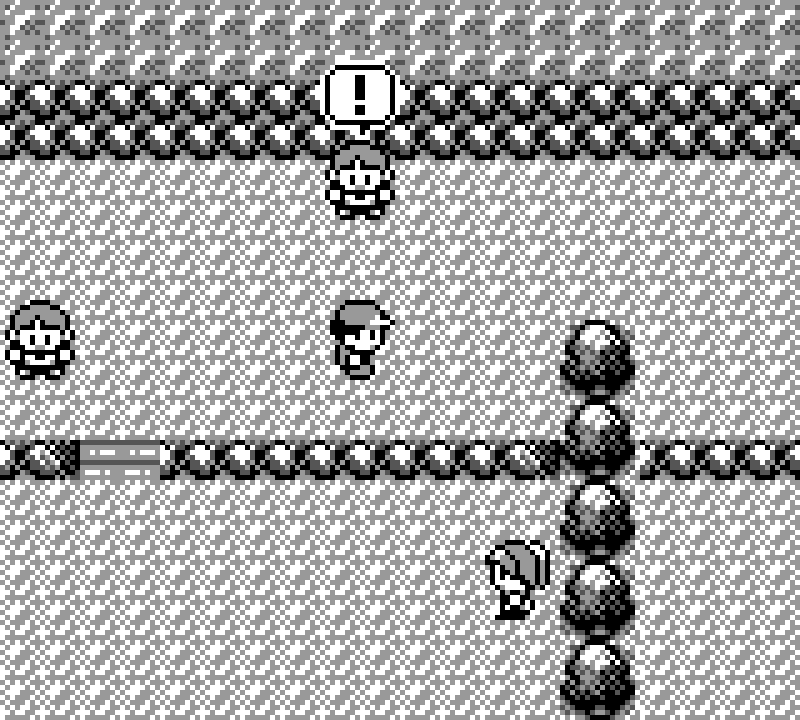}
        \caption{Mandatory Trainer Battle Triggered on Line-of-sight}
        \label{fig:route3}
    \end{subfigure}%
    \hfill
    \begin{subfigure}[b]{0.24\textwidth}
        \includegraphics[width=\textwidth]{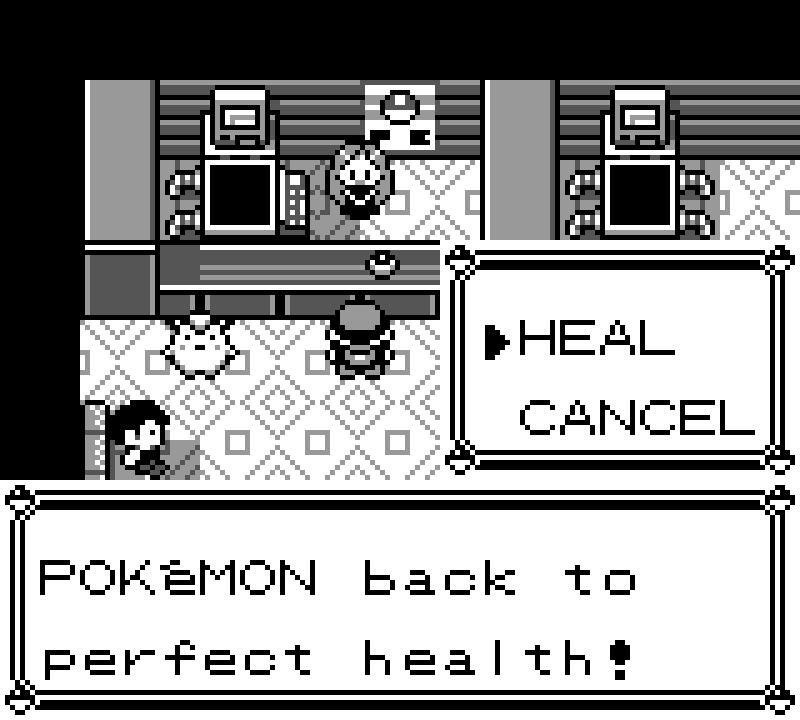}
        \caption{Pokémon Center Healing by Interacting with Nurse Joy}
        \label{fig:poke_center_heal}
    \end{subfigure}%
    \hfill
    \begin{subfigure}[b]{0.24\textwidth}
        \includegraphics[width=\textwidth]{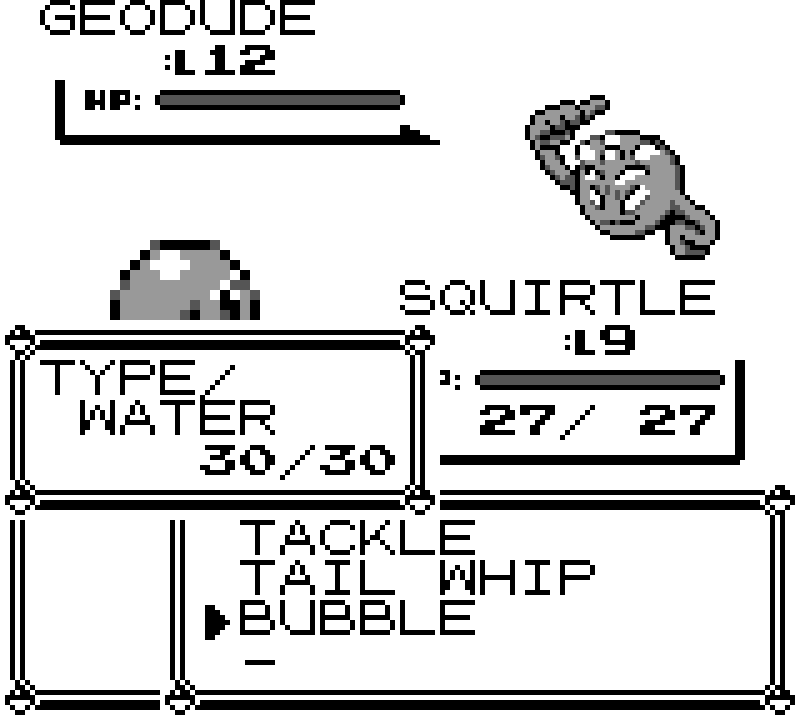}
        \caption{Pokémon Battle: Selecting a Super Effective Attack}
        \label{fig:battle_geodude_2}
    \end{subfigure}%
    
    \vspace{1em} 

    \begin{subfigure}[b]{0.24\textwidth}
        \includegraphics[width=\textwidth]{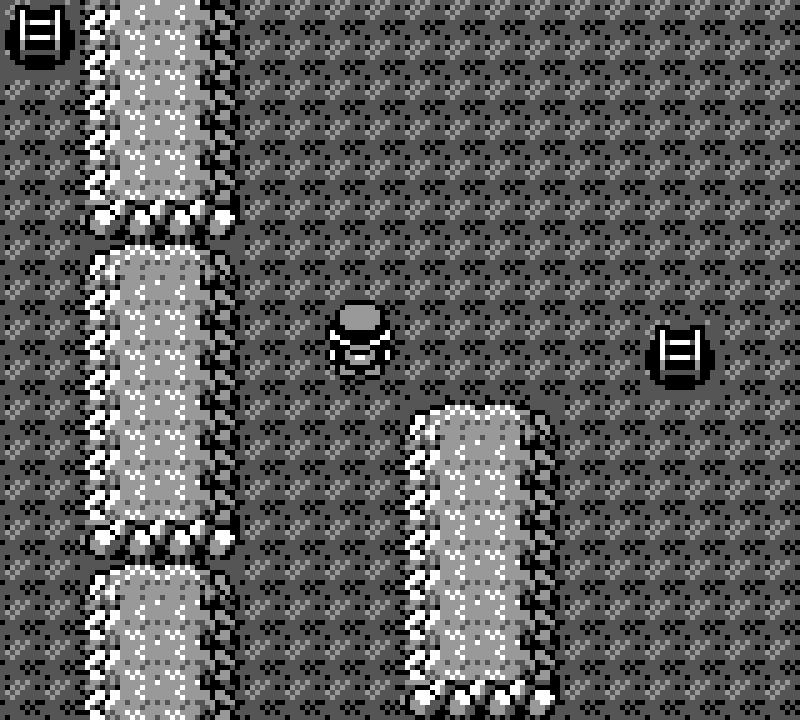}
        \caption{Mt. Moon: Maze-like Dungeon Full of Wild Pokémon}
        \label{fig:mt_moon}
    \end{subfigure}%
    \hfill
    \begin{subfigure}[b]{0.24\textwidth}
        \includegraphics[width=\textwidth]{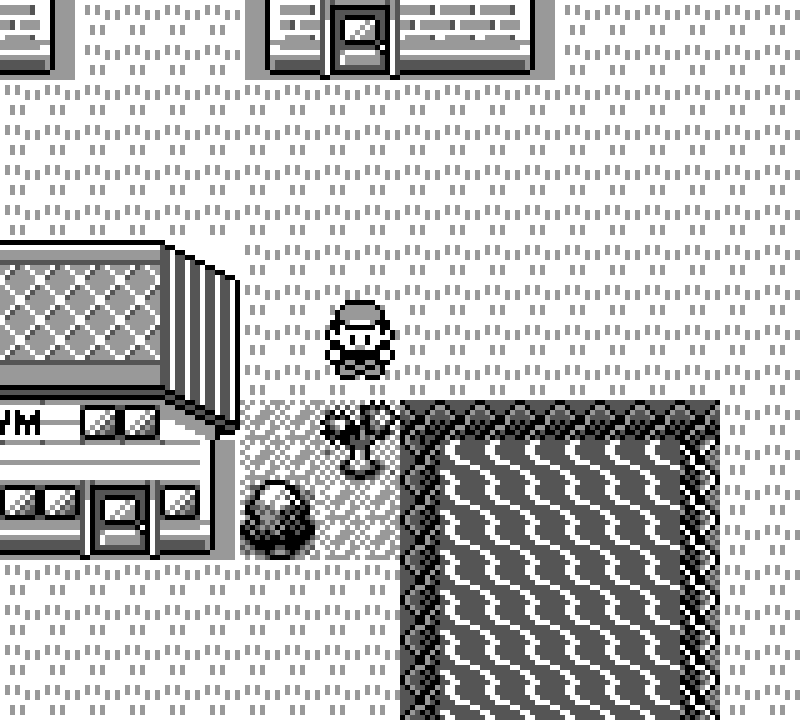}
        \caption{Cuttable Tree as Obstacle and Game Mechanic}
        \label{fig:vermilion_cut}
    \end{subfigure}%
    \hfill    
    \begin{subfigure}[b]{0.24\textwidth}
        \includegraphics[width=\textwidth]{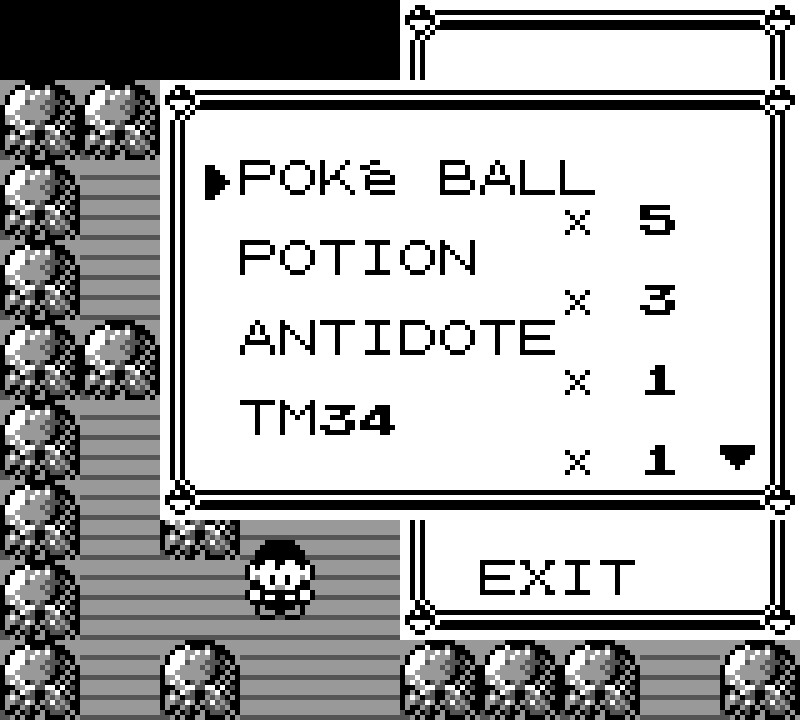}
        \caption{Item Menu outside of Battle\\~}
        \label{fig:item_menu}
    \end{subfigure}%
    \hfill
    \begin{subfigure}[b]{0.24\textwidth}
        \includegraphics[width=\textwidth]{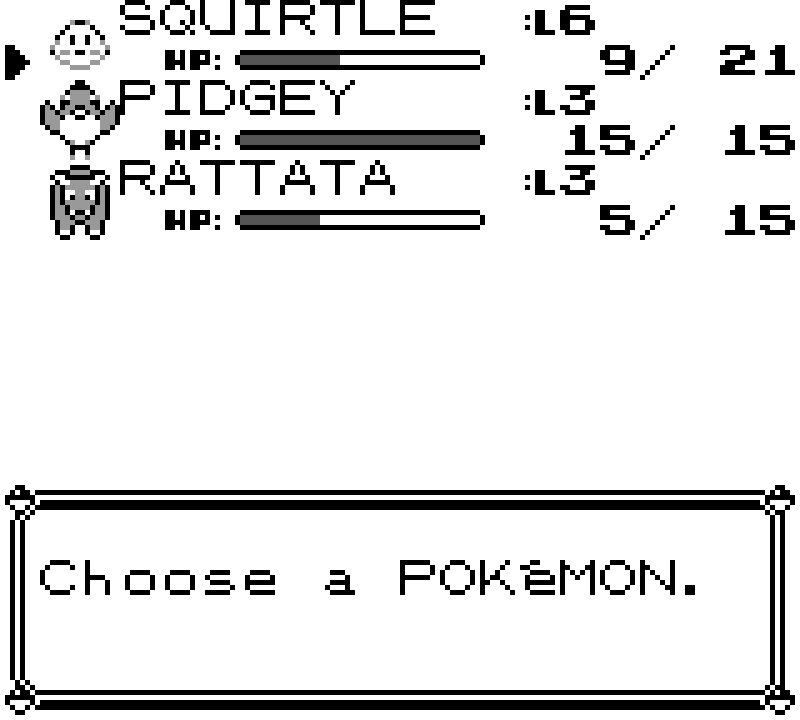}
        \caption{Party Menu Displays Party Pokémon Order, HP, and Level}
        \label{fig:party_menu_2}
    \end{subfigure}%
    \caption{All subfigures depict game screens possibly perceivable by the agent, presenting various challenges in Pokémon Red that involve exploration, navigation, and strategic decision-making. The agent must traverse a 2D overworld (\subref{fig:route1}) with a party of Pokémon (\subref{fig:party_menu_2}), winning mandatory trainer battles (\subref{fig:route3}) and navigating complex maze-like areas (\subref{fig:mt_moon}). Progression also depends on interacting with the game's UI, including healing Pokémon at a Pokémon Center (\subref{fig:poke_center_heal}), using strategy to win battles (\subref{fig:battle_geodude_2}), using items (\subref{fig:item_menu}). Overcoming obstacles, like cuttable trees (\subref{fig:vermilion_cut}), requires the use of mandatory game-mechanic moves, which must be obtained via exploration, taught to eligible Pokémon, and used via the Start menu interface when facing the obstacle.}

    \label{fig:gameplay}
\end{figure*}

Video games have significantly advanced artificial intelligence, acting as both a catalyst for research and a benchmark for solving complex problems.
DRL and its adjacent fields have achieved remarkable results by enabling the training of agents to play games with extremely long horizons such as DotA~2 \cite{Berner2019Dota2}, NetHack \cite{kuettler2020nethack}, and Minecraft \cite{minedojo2022}. While these modern games were used to demonstrate strong advancements, older games remain an untapped resource, offering similar levels of complexity, unique challenges, and opportunities for research.
This work examines an older game, the 1996 Game Boy JRPG: Pokémon Red.

Pokémon games have consistently gained widespread attention, both within and beyond the gaming community.
A 2023 YouTube video showcasing an early DRL agent playing Pokémon Red went viral, garnering over 7.5 million views \cite{pdubsYoutube}.
The video's viral success highlights the game's broad appeal and serves as the inspiration that we build upon. 
Pokémon and its many successors challenge the player with a variety of tasks with the ultimate goal of becoming a Pokémon Master by defeating the game's Champion and catching all 151 Pokémon species.
In this paper, we highlight Pokémon Red as a game that exposes players to multiple complex tasks:
\begin{itemize}
\item Multi-Task Challenge: Pokémon includes strategic battling, 2D navigation, UI navigation, resource management, and puzzle-solving (Figure \ref{fig:gameplay}).
\item Exploration Challenge: 
Beating Pokémon requires careful planning and navigation. In our experiments, the trained DRL agents made tens of thousands of decisions per episode to complete only a small fraction of the game.
\item Large Policy Space: With 151 Pokémon species, diverse combinations of moves, and numerous usable items, Pokémon Red features vast gameplay possibilities and freedom to develop successful policies. 
\end{itemize}

To explore Pokémon Red’s potential, we contribute a minimal DRL environment and a simple, adaptable baseline agent trained with Proximal Policy Optimization (PPO) \cite{Schulman2017}.
Our approach lays the groundwork for more advanced exploration and hierarchical methods, while offering detailed insights into the performance and behavior of the trained baseline agent and various reward ablations.
This paper is structured as follows: we establish the Pokémon baseline by introducing the gameplay and properties of the Pokémon Red environment; then, the training setup is detailed.
Section \ref{sec:experiments} presents results, discussing the key findings and potential future directions.
Before concluding, we refer to related work.

\section{\textbf{Pokémon Red Environment}}

This section describes the core mechanics of Pokémon Red and our formulation as a Markov Decision Process (MDP) \cite{sutton2018reinforcement}, including the design of the observation space, action space, reward function, and terminal conditions.
The simulation speed of the environment was measured using random actions on an AMD Ryzen 7 2700X CPU. Pokémon runs at $\approx$9403 steps per second (SPS), which is faster than Atari Breakout (7117 SPS) and slower than Procgen (18530 SPS) \cite{pleines2023memory}.

\subsection{Selected Game Objectives}

In this work, we will focus on Pokémon Red's storyline objectives, which entail completing several quests and milestones.
For simplicity, we have reduced the objective from completing the entire game to completing only the main tasks up to the end of Cerulean City, the location of the second gym.
This constitutes approximately 20\% of the full game and is achieved by accomplishing the following milestones:

\begin{itemize}
\item Start in Pallet Town and pick a starter Pokémon from Professor Oak.
\item Travel to Viridian City and obtain Oak's Parcel.
\item Return the parcel to Oak's lab in Pallet Town.
\item Traverse Viridian Forest to reach Pewter City.
\item Defeat Pewter City's gym leader, Brock (Badge 1).
\item Traverse Mt. Moon to arrive at Cerulean City.
\item Defeat Cerulean City's gym leader, Misty (Badge 2).
\item Transform Bill, a storyline non-playable character (NPC), back into a human to unlock the path to Vermilion City.
\end{itemize}

Because the first two objectives require backtracking, we have started the environment after the third objective, as the remaining tasks involve minimal or no backtracking.
Thus, the agent does not choose its starter.
Instead, we let it start with the water-type Pokémon Squirtle, which has an advantage against gym leader Brock's rock Pokémon.
After traversing Mt. Moon, the subsequent storyline objectives can be completed in a variety of orders, and the game transitions into a large, open-world setting with numerous non-linear tasks and puzzles.

\subsection{Game Mechanics}

Pokémon Red features turn-based combat, strategic team-building, and exploration of an expansive grid world.
The player navigates interconnected maps, overcoming obstacles such as trainer battles, environmental barriers, and puzzles.
Early on, battles are the primary challenge.

Battles initiate with the first Pokémon in each player's party (a wild Pokémon is treated as a party of one).
Battles are turn-based, with each side selecting moves that are executed in order based on the Pokémon's Speed stat, so that faster Pokémon act before slower ones.
On each turn, a player may choose a move (Fig.~\ref{fig:battle_geodude_2}), switch to another Pokémon in the party (Fig.~\ref{fig:party_menu_2}), use an item, or run from battle.
Escaping from battle is only possible during wild encounters.
Wild Pokémon appear randomly in grass, on water, or inside dungeons, and players can capture them using Poké Balls.
Each Pokémon can have up to four moves, each with a finite number of Power Points (PP) representing the number of times a move can be used in battle.
Using a move consumes one PP and may heal, deal damage, inflict status effects, or temporarily alter stats.
The effects of a move can vary based on the attacking and target Pokémon's types.
This type match-up system is analogous to rock-paper-scissors, where each type holds a natural advantage over another.
For example, water-type moves deal double damage against fire-type Pokémon.
Attacks can also randomly become Critical Hits that deal double damage.
When a Pokémon's Hit Points (HP) reach zero, it faints and can no longer be used in battle.
A Pokémon's HP and PP can be restored using items (Fig.~\ref{fig:item_menu}) or by visiting a Pokémon Center (Fig.~\ref{fig:poke_center_heal}).
Defeating a wild or trainer Pokémon grants experience points (XP) to all conscious Pokémon that participated in battle.
To gain XP, a Pokémon must remain conscious until the enemy Pokémon faints.
Gaining XP leads to level-ups that increase a Pokémon's stats and offer opportunities to learn new moves or evolve.
Evolution is triggered at certain levels and can be canceled with the B button; it raises a Pokémon's stats but may delay learning new moves, offering a strategic tradeoff.
Many Pokémon have evolved forms, and some moves are learned at lower levels in their base forms than after evolving.
If the player's entire party faints, it is considered a "blackout," and the player will respawn at the last visited Pokémon Center or at their home in Pallet Town if no Pokémon Centers have been visited.

\subsection{Observation Space}

The agent receives two observation modalities: a vision input and a game state vector.

The vision input uses a grayscale representation of the Game Boy's screen downsampled by two for a resolution of 72×80 pixels.
We stack the current frame and two previously observed frames to provide minimal temporal awareness.
The agent additionally observes a separate 48×48 binary crop of the screen, centered on the player, that indicates visited coordinates.

The game state vector includes the current HP and level of each Pokémon in the party, as well as flags indicating the completion status of various in-game events.
Notably, this observation space is intentionally limited for simplicity and to mimick the information a first time human player would have.
The observation does not provide information on the species, stats, nor movesets of the Pokémon.

\subsection{Action Space}

The action space is discrete and consists of seven Game Boy buttons: A, B, Start, Up, Down, Left, and Right (Select is omitted).
An agent must select one of these actions for each environment step.
Navigation of both the overworld and in-game menus is performed using the arrow keys.
The A button confirms selections or initiates an interaction, while the B button cancels actions or exits menus.
The Start button opens the main menu.

For simplicity, actions are represented as full presses instead of having a press/unpress action for each button.
To reliably move one tile at a time in the grid world, each button is held for 8 frames, then released for 16 frames, resulting in one decision every 24 frames.
As a result, the simulation runs at a slower speed of approximately $\frac{9403}{24} \approx 392,\text{SPS}$.

\subsection{Determinism and Terminal Conditions}

Wild encounters, Pokémon stats, critical hits, and other mechanics appear stochastic.  
However, the random number generator (RNG) on the Game Boy is influenced by the player's input.  
As a result, if the player behaves deterministically, the environment will also behave deterministically. 
This property is commonly exploited by speedrunners and could potentially be leveraged by trained agents as well \cite{pdubsYoutube}.  
The agent could execute a specific sequence of inputs to encounter a strong Pokémon and catch it on the first try.
To counteract this determinism, a sequence of random button presses could be executed upon resetting the environment.

Regarding terminal conditions, time is the only constraint we define.
The agent begins with a budget of 10,240 steps, which increases by 2,048 steps for each completed event.  
Events include important trainer battles, distinct NPC interactions, and storyline progression.  
This dynamic step budget introduces variety in training samples, preventing environments from resetting at fixed intervals and mitigating the risk of catastrophic forgetting (Section \ref{sec:horizons}).

\subsection{Reward Function}

Pokémon has an inherently long time horizon, requiring hours of real-world play time before major objectives are achieved. We use dense auxiliary rewards to reinforce intermediate progress towards and behaviors that support reaching these goals.

\textbf{Event Reward}: The agent receives a reward of $ R_{\text{event}} $ = +2 for every completed event (trainer battle, step in quest progression).

\textbf{Navigation Reward}: Without a navigation reward, the agent converges to a policy that does not progress much further than the player's origin in Pallet Town. To aid in exploring the overworld, the agent receives $ R_{\text{nav}} $= +0.005 for each new overworld coordinate visited within an episode.

\textbf{Healing Reward}: When HP is gained, the agent receives a reward proportional to the fractional sum of party HP:  
\begin{equation}
    R_{\text{heal}} = 2.5\sum_{i=1}^{6} \frac{\text{HP}_i^{\text{after}} - \text{HP}_i^{\text{before}}}{\text{HP}_i^{\text{max}}}
\end{equation}  
This reward is triggered when a Pokémon levels up, when a Pokémon Center is used, or when a potion is used on a Pokémon with less than full HP; each of these behaviors supports progression.

\textbf{Level Reward}: The agent receives a reward based on the levels the Pokémon in its party. This encourages early level progression and the capture of new Pokémon while discouraging wild-encounter grinding.
To balance this behavior, we define:
\begin{equation}
    R_{\text{lvl}} = 
    0.5\min\left(\sum_{i=1}^{6} \text{lvl}_i, \frac{\sum_{i=1}^{6} \text{lvl}_i - 22}{4} + 22 \right)
\end{equation}
The threshold of 22 signifies a reasonable level to battle the leader of Gym 2, Misty, in Cerulean City, with the marginal gain per additional level downscaled by a factor of 4.

Finally, all rewards are summed up to yield \( R = R_{\text{event}} + R_{\text{nav}} + R_{\text{heal}} + R_{\text{lvl}} \). Rewards are further discussed in Section \ref{sec:reward_shaping}.

\section{Training Method}

All agents are trained using Proximal Policy Optimization (PPO) \cite{Schulman2017}.
We leverage the clipped surrogate loss to optimize the policy at time $t$:
\begin{equation}
\begin{aligned}
    L^{C}_t(\theta) = -\mathbb{E}_t \Bigg[ &\min \Big( q_t(\theta) A_\pi^{\text{GAE}}(s_t, a_t), \\
    &\quad \text{clip}(q_t(\theta),1- \epsilon,1+\epsilon) A_\pi^{\text{GAE}}(s_t, a_t) \Big) \Bigg]
\end{aligned}
\label{eq:policy_loss}
\end{equation}
\begin{equation*}
    \text{where} \quad q_t(\theta) = \frac{\pi(a_t|s_t,\theta)}{\pi(a_t|s_t, \theta_{\text{old}})}
\end{equation*}
$s_t$ denotes the observed state, $a_t$ is the chosen action, $\theta$ defines the parameters of the policy, $\epsilon$ is the clip range, and $A_\pi^{\text{GAE}}$ is the generalized advantage estimate.
The value function shares parameters with the policy and undergoes clipping:
\begin{equation}
\begin{aligned}
    L^{VClip}_t(\theta) = \Big( &\text{clip} \big( V(s_t, \theta), \\
    &V(s_t, \theta_{\text{old}}) - \epsilon, \\
    &V(s_t, \theta_{\text{old}}) + \epsilon \big) - \hat{V}_t \Big)^2
\end{aligned}
\end{equation}
\begin{equation}
    L_t^{V}(\theta) = \max\left[\left(V(s_t, \theta) - \hat{V}_t\right)^2,~L^{VClip}_t(\theta)\right]
    \label{eq:value_clip}
\end{equation}
The final combined loss is depicted by $L^{C+V}_t(\theta)$:

\begin{equation}
	L^{C+V}_t(\theta)=\mathbb{E}_t [L^{C}_t(\theta)+vL^{V}_t(\theta)]
\end{equation}
\noindent
where $v$ is the value function's coefficient.

We intentionally leave out the commonly used entropy bonus, as tuning its weighting across \(10^{-3}\), \(10^{-4}\), \(10^{-5}\), and \(10^{-6}\) yielded no significant improvements.

\subsection{Network Architecture}

\begin{figure}
    \centering
    \includegraphics[width=0.75\linewidth]{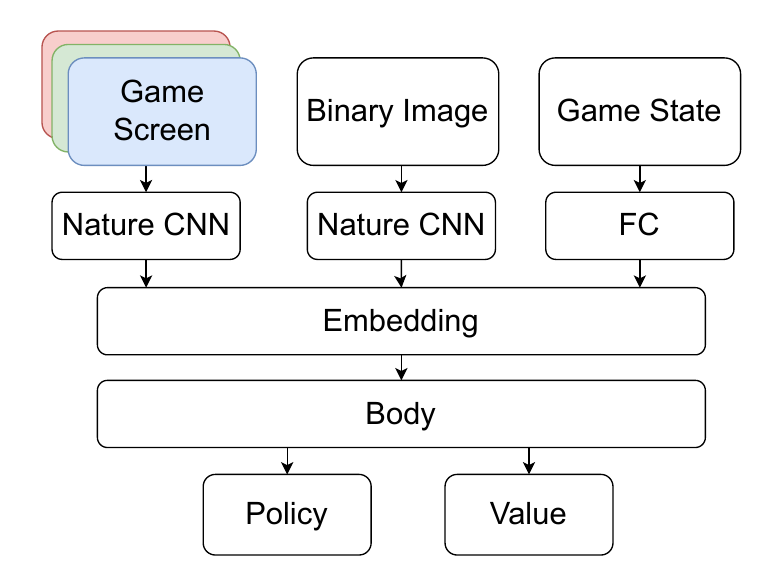}
    \caption{Actor-Critic Network (2M parameters, GRU: 4M).}
    \label{fig:ann}
\end{figure}

The general network architecture of the agent is illustrated in Figure \ref{fig:ann}.
Each vision modality within the observation space is encoded using a classic Nature Convolutional Neural Network (CNN) \cite{mnih2015DQN}.
The game state vector is processed separately through a single fully connected layer.
The outputs of all encoders are then flattened, concatenated, and projected to match the dimensionality of the network’s body, which consists of a fully connected layer as standard, or, to facilitate recurrence, a Gated Recurrent Unit (GRU).
Pleines et al. \cite{pleines2023memory} describe how the GRU enables memory capabilities by maintaining a hidden state over time.
The hidden features computed in the body are used to construct both a policy head and a value head, which respectively output the discrete policy (actions) and state value estimate (via the value function).
The ReLU activation function is applied throughout the network.

\subsection{Hyperparameters}

\begin{table}
\centering
\caption{The hyperparameters used in our experiments.}
\begin{tabular}{l r l r}
\toprule
Hyperparameter            & Value  & Hyperparameter             & Value  \\
\midrule
Discount Factor $\gamma$  & 0.997  & Clip Range $\epsilon$      & 0.2    \\
$\lambda$ (GAE)           & 0.95   & Value Function Coef. $v$   & 0.5    \\
Number of Workers         & 32     & Learning Rate              & 0.0003 \\
Worker Steps              & 2048   & Max Gradient Norm          & 0.5    \\
Epochs                    & 3      & Activations                & ReLU   \\
Mini Batches              & 8      & Optimizer                  & AdamW  \\
\bottomrule
\end{tabular}
\label{tab:hyperparameters}
\end{table}

Table \ref{tab:hyperparameters} presents the hyperparameters used across all experiments.
Since Pokémon Red features episodes lasting tens of thousands of steps, each environment instance (i.e., worker) samples a horizon of 2048 steps, which we found to be more effective than horizons of 512 or 1024 steps.
This longer horizon increases the likelihood of capturing more reward signals compared to the more commonly used shorter horizons, such as 512 steps or fewer.
To leverage all available cores of our training hardware, we use 32 workers, resulting in a total batch size of 65,536.
\(\gamma\) was tuned over \(\{0.99, 0.997, 0.999, 0.9995\}\).
\section{Experimental Results and Discussion}
\label{sec:experiments}

The previous sections introduced the MDP behind Pokémon Red and the training algorithm.  
These establish our minimal baseline, which we analyze through ablation experiments.  
We vary the agent's starting state by selecting Squirtle, Charmander, or Bulbasaur as the starter Pokémon.  
Additional experiments ablate individual reward signals and compare the results to 30 playthroughs by skilled human players.

Each experiment is repeated five times with different seeds.  
During training, we evaluate 25 time points using 30 episodes per repetition.  
We report the mean performance across runs and measure variability with the standard deviation.  

\begin{figure*}[t] 
    \centering

    \begin{subfigure}[t]{0.8\textwidth}
        \centering
        \includegraphics[width=\textwidth, trim={0 20 0 26}, clip]{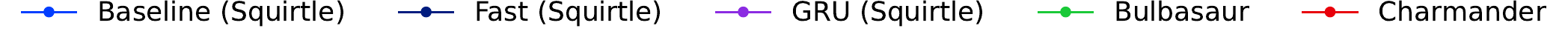}
    \end{subfigure}

    \begin{subfigure}[t]{0.9\textwidth}
        \centering
        \includegraphics[width=\textwidth, trim={0 42 0 0}, clip]{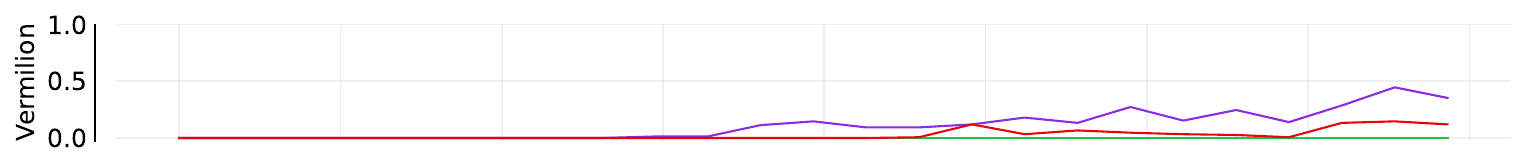}
    \end{subfigure}

    \vspace{0.25em} 

    \begin{subfigure}[t]{0.9\textwidth}
        \centering
        \includegraphics[width=\textwidth, trim={0 42 0 0}, clip]{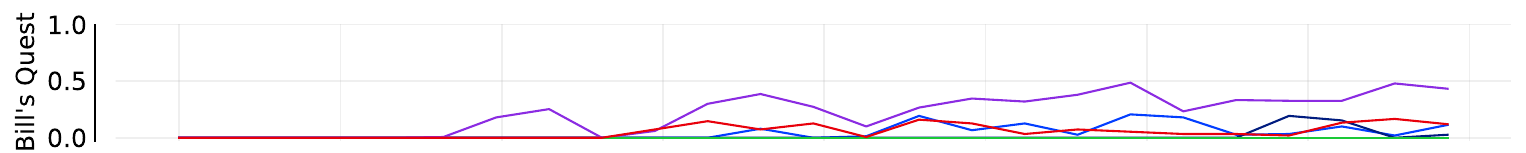}
    \end{subfigure}

    \vspace{0.25em}

    \begin{subfigure}[t]{0.9\textwidth}
        \centering
        \includegraphics[width=\textwidth, trim={0 42 0 0}, clip]{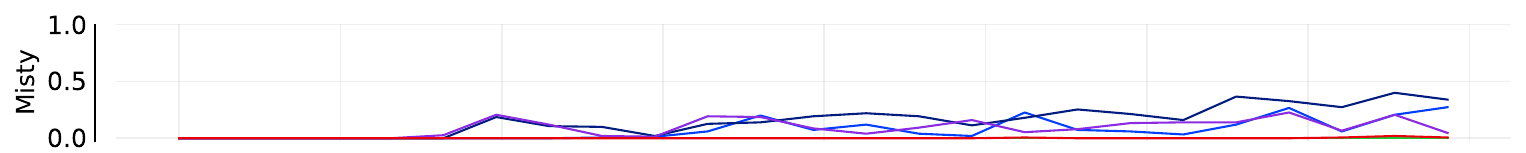}
    \end{subfigure}

    \vspace{0.25em}

    \begin{subfigure}[t]{0.9\textwidth}
        \centering
        \includegraphics[width=\textwidth, trim={0 42 0 0}, clip]{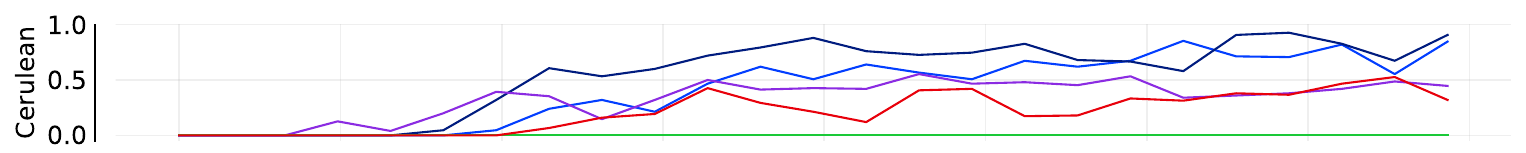}
    \end{subfigure}

    \vspace{0.25em}

    \begin{subfigure}[t]{0.9\textwidth}
        \centering
        \includegraphics[width=\textwidth, trim={0 0 0 0}, clip]{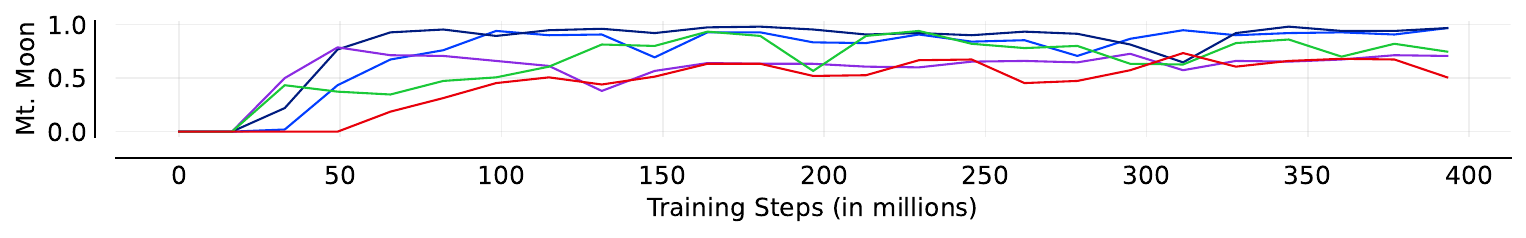}
    \end{subfigure}

    \caption{Mean completion curves for distinct milestones. Beating Misty can be skipped to reach Vermilion City.}
    \label{fig:performance_overview}
\end{figure*}

\subsection{Baseline Performance}

Figure \ref{fig:performance_overview} shows sample efficiency curves, depicting the mean proportion of milestones completed.
The experiments vary by starter Pokémon, an agent using memory via GRU, and a \textit{Fast} setting with increased text speed and disabled battle animations.

The first milestone is reaching Mt. Moon after defeating Brock.
\textit{Fast} performs best, peaking at 98\% completion rate, followed by the baseline at 97\%.
With Bulbasaur, the agent reaches 94\% but fails entirely at the second milestone: arriving in Cerulean City.
This outcome is due to exploiting the heal reward as discussed in Section \ref{sec:heal_exploit}.
Other agents achieve between 53\% and 93\%.

Beating Misty and completing Bill's quest can be done in any order, but only the latter unlocks Vermilion City.
Agents may skip Misty since her water-type Pokémon pose a challenge, especially for the fire-type Charmander, which has just a 2\% completion rate.
\textit{Fast} (33\%), baseline (27\%), and GRU (21\%) perform better, but the rates are low compared to reaching Cerulean City.
While \textit{Fast} and baseline excel early on, the GRU agent performs better on Bill’s quest and on reaching Vermilion City.  
Bill's quest is not trivial, requiring three basic interactions, in a specified order, without leaving his house. The events reset if the house is exited before completion of the full sequence. Despite the confined space of the house, and an event flags observation that should convey quest progression to agents, completion still proves too complex for most:  
for example, 71\% of \textit{Fast} experiments visited Bill's house, yet only 19\% of all \textit{Fast} completed his quest.
In contrast, 48\% of GRU agents completed the quest, down just 1\% from the 49\% arriving at Cerulean; this suggests an advantage in retaining task-relevant memory.

\subsection{Discussing Ablations and Human Performance}
\label{sec:ablations}
Table \ref{tab:performance} adds the following experiments to those previously described: individual rewards ablations, starting the agent from a state where it can choose its starter Pokémon, and the results from human playthroughs through Cerulean City.
Next, we discuss the most salient findings.

\subsubsection{Distinct Exploration Strategies are Indispensable}
\label{sec:ablations}

Exploring the world of Pokémon is tremendously difficult.  
To address this, a naive navigation reward is introduced.  
Without it, the agent fails to achieve any milestones, yet, when this reward is scaled up by a factor of 10, the agent rarely defeats Brock. Instead, such agents explore almost exclusively, to the detriment of all other aspects of the game. Even battles and menuing are avoided to conserve steps and thus maximize navigation reward.

Since this reward signal is so simplistic, yet navigation is so crucial, it would be interesting to investigate more advanced exploration methods, such as curiosity-driven rewards \cite{Pathak2017} or Random Network Distillation \cite{Burda2019}. Such methods could provide robustness via reduced reward sensitivity, and by potentially allowing the visited-coordinates binary observation to be obviated.

\subsubsection{Coping with Extremely Long Horizons}
\label{sec:horizons}

Most MDPs have a terminal state, but in Pokémon Red, we must define this ourselves due to the game's open-ended nature.
With a fixed episode length, all data-collecting workers reset simultaneously, repeatedly sampling similar training data.
Since the game's episodes are significantly longer than the sampled trajectories, there is a high risk of catastrophic forgetting, where learned policies are lost.
To mitigate forgetting, we allowed the episode length to grow dynamically based on progress.
Because the agent's decisions are stochastic, workers quickly fall out of sync in their resets, with some agents restarting in Pallet Town, while others continue advancing. While this increases sample diversity, training remains unstable: if Figure \ref{fig:performance_overview} rendered standard deviations, the performances across different run types would be difficult to distinguish.  
Similarly, Table \ref{tab:performance} hints on this instability, showing large standard deviations.  

Although agents are competitive in defeating Brock, they fall significantly behind in reaching Cerulean City: humans require about 11,000 steps on average to reach Cerulean City, while an agent requires effectively double that. The \textit{Fast} agent stood out among agents as the quickest, requiring just 18,500 steps on average. Since \textit{Fast} greatly speeds the passage of the dialogs that appear upon NPC or object interaction, and when in battle, and clearing such dialogs requires multiple successive, correct inputs, it is likely that action efficiency plays an important role in completion speed. A short dialog takes at least 10 steps; evolution, another dialog, takes about 15 steps. Here, steps can be saved by pressing B, which aborts the evolution. Although human players can use this advantageously, agents only seem to stochastically abort, suggesting an insouciance towards wasted steps.
Since addressing such inefficiency would require additional engineering, such as reading memory locations of the game to determine when decision-making is relevant, while our action execution method is precise for overworld navigation, it remains coarse for battles and dialogues.
One approach to handle unnecessary actions could be to add a time dimension to the action space.  
In addition to selecting a button to press, the agent would decide when to act or how long to keep the button pressed.  
This would allow the agent to learn when to act and when to observe by itself \cite{zhou2024}.

Finally, the suitability of the training method itself is questionable.
Maximizing the discounted cumulative reward encourages minimizing time, introducing bias.
This bias is evident in the \textit{Choose Starter} experiment, where the agent consistently picks Charmander, despite Squirtle being the most effective choice against Brock.
The agent favors Charmander because Squirtle and Bulbasaur require one and two extra actions, respectively, to reach in the overworld compared to Charmander.
Since picking a starter immediately grants an event and level reward, Charmander is chosen simply because its reward is triggered sooner.
By the time the agent reaches Brock, it becomes clear that Charmander is an inferior choice; however, defeating Brock takes over 3,000 steps on average—vastly exceeding both the sampled trajectory length of 2,048 steps and the effective horizon of approximately $333$ steps (calculated as $\frac{1}{1-0.997}$).
To address this, hierarchical training methods, such as Director \cite{hafner2022deep}, are promising.
In this context, a sequence of primitive actions (i.e., button presses) can be combined to form high-level actions.
Choosing the starter Pokémon could be such an action, which would remove the undesirable time bias.
However, learning high-level actions from scratch remains a significant challenge.

\subsubsection{Heal Reward Exploitation}
\label{sec:heal_exploit}

The heal reward encourages exploitation, especially when Bulbasaur is chosen as the starter.
The agent's policy converges to a strategy of battling wild Zubats in Mt. Moon, using Bulbasaur's Leech Seed to restore health while Zubat heals with Leech Life.
This results in nearly endless battles, providing frequent heal rewards.
Before Bulbasaur faints, the agent visits the Poké Center for another heal reward.

Similar exploitation occurs in the Charmander, GRU, and \textit{Choose Starter} experiments, where at least one out of five runs involves battling wild Pokémon near Pallet Town and using the player's mother to heal.
This strategy leads to suboptimal performance, with runs failing to exceed an 80\% success rate against Brock (Table \ref{tab:performance}) and high standard deviations in heals and Poké Center visits (Table \ref{tab:policy}).
Removing the heal reward generally reduces performance, as the Poké Center is no longer used as a respawn point.
With an average of 8–9 blackouts, the agent requires additional steps to retrace its path from Pallet Town to the point of fainting (e.g., Cerulean City).

Bulbasaur is an exception, as the agent reaches Cerulean City without the heal reward, though it is less effective than Charmander or Squirtle.
Bulbasaur may appear advantageous due to its grass-type advantage against Brock, but Bulbasaur learns Vine Whip, its first contact Grass-type attack, comparatively late, at level 15. And, this move has a pitiful 10 PP.
In contrast, Squirtle learns Bubble at level 8, and Charmander learns Ember at level 9; both moves have higher PP than Vine Whip, making them more suitable for the lengthy, battle-heavy traversal of Mt. Moon. But such nuance is lost on the agent: the starter Pokémon is chosen very early in the episode, and connecting this choice to distal outcomes would require significant engineering.

\subsubsection{Reward Shaping Introduces Vulnerabilities}
\label{sec:reward_shaping}

Beyond healing and navigation, event and level rewards also shape the agent’s behavior in unintended ways.

Event rewards serve as breadcrumbs, guiding the agent through the storyline.  
Some, like retrieving a potion from an NPC, are optional, while others, such as earning gym badges, are mandatory.  
That said, the agent frequently skips Misty's badge in Cerulean City while progressing to Vermilion City, suggesting that critical events may require stronger feedback.  

Ablating the level reward yields the best ablation performance (Table \ref{tab:performance}), but introduces side effects as well.  
The agent’s highest-level Pokémon averages 27.45—suprisingly comparable to the baseline, and just two levels below the \textit{Fast} run.  
To explain this performance, we hypothesize that ablating level reward might allow the agent to `focus' on other sources of rewards, namely, event rewards. Since most events are trainer battles, which are early-game dense and XP-rich, this agent progresses readily within the scope of this paper by having sufficient level and incentive to readily complete the early game objectives. 
    While all reward sources influence the agent's performance, other rewards have a debatably-positive effect on the agent. 

An agent starting with Squirtle (baseline) typically avoids healing exploits, whereas the Charmander agent heals heavily. But, ablate the heal reward, and the agent explores fecklessly, visiting 5637 coordinates on average. Despite varying positive and side effects, no reward nor ablation manages to address the challenge of reaching the third gym in Vermilion City, which is blocked by a cuttable tree (Figure \ref{fig:vermilion_cut}).
This presents a significant exploration challenge:  
the agent must obtain a Pokémon capable of learning Cut, acquire the 
necessary item, HM01, navigate the UI to use it, and correctly execute the move in the overworld.
Until to this point in the game, none of these actions were necessary. Clearly, solving this requires
\begin{landscape}  

    \begin{table}[p]
        \centering
        \caption{Performance results of various ablation experiments. For each metric, the best data point—determined by milestone completion over time—is selected. Rows are sorted in descending order by the number of completed milestones. The tracked milestones include reaching Viridian Forest, defeating Brock, reaching Mt. Moon, arriving at Cerulean City, defeating Misty, solving Bill's quest, and completing Cerulean City by defeating the Rocket Thief. The best values in each column are highlighted in bold. The \textit{Fast} experiment saves time (i.e., steps) by setting the game's text speed to fast and disabling battle animations. The \textit{Choose Starter} experiment does not skip the Oak parcel quest at the beginning of the game, instead starting at the beginning where the player still needs to selects their first Pokémon.}

        \resizebox{\linewidth}{!}{  

\begin{tabular}{lllllllllll}
        \toprule
        Experiment & Milestones & Beat Brock & Mt Moon & Cerulean City & Beat Misty & Bill Quest & Cerulean Done & Beat Brock Steps & Cerulean City Steps & Cerulean Done Steps \\
        \midrule
        Human & $7.00 \pm 0.0$ & $1.00 \pm 0.0$ & $1.00 \pm 0.0$ & $1.00 \pm 0.0$ & $1.00 \pm 0.0$ & $1.00 \pm 0.0$ & $1.00 \pm 0.0$ & $5403 \pm 3824$ & $11188 \pm 5224$ & $16842 \pm 7413$ \\ \hline
        Baseline${~-~R_{lvl}}$ & $\mathbf{4.60 \pm 0.7}$ & $0.95 \pm 0.0$ & $0.91 \pm 0.1$ & $0.79 \pm 0.1$ & $0.31 \pm 0.3$ & $0.32 \pm 0.4$ & $0.03 \pm 0.1$ & $6117 \pm 1758$ & $21352 \pm 3898$ & $\mathbf{37938 \pm 0}$ \\
        Fast & $4.40 \pm 0.8$ & $0.98 \pm 0.0$ & $\mathbf{0.98 \pm 0.0}$ & $\mathbf{0.93 \pm 0.1}$ & $\mathbf{0.33 \pm 0.4}$ & $0.19 \pm 0.4$ & $0.00 \pm 0.0$ & $\mathbf{3753 \pm 990}$ & $\mathbf{18523 \pm 4280}$ & -- \\
        Baseline & $4.19 \pm 0.2$ & $\mathbf{0.99 \pm 0.0}$ & $0.97 \pm 0.0$ & $0.85 \pm 0.1$ & $0.27 \pm 0.3$ & $0.11 \pm 0.2$ & $0.00 \pm 0.0$ & $5587 \pm 1235$ & $25299 \pm 4950$ & -- \\
        Baseline${~-~R_{heal}}$ & $3.94 \pm 0.2$ & $\mathbf{0.99 \pm 0.0}$ & $0.97 \pm 0.0$ & $0.91 \pm 0.1$ & $0.07 \pm 0.1$ & $0.00 \pm 0.0$ & $0.00 \pm 0.0$ & $5497 \pm 503$ & $28813 \pm 2880$ & -- \\
        GRU & $3.93 \pm 2.2$ & $0.76 \pm 0.4$ & $0.71 \pm 0.4$ & $0.49 \pm 0.4$ & $0.21 \pm 0.3$ & $\mathbf{0.48 \pm 0.4}$ & $\mathbf{0.21 \pm 0.3}$ & $5624 \pm 780$ & $27118 \pm 4236$ & $46532 \pm 2249$ \\
        Fast${-R_{heal}}$ & $3.77 \pm 0.5$ & $0.97 \pm 0.0$ & $0.96 \pm 0.0$ & $0.73 \pm 0.3$ & $0.12 \pm 0.2$ & $0.00 \pm 0.0$ & $0.00 \pm 0.0$ & $6531 \pm 1185$ & $28871 \pm 1362$ & -- \\
        Charmander${~-~R_{heal}}$ & $3.73 \pm 0.2$ & $0.94 \pm 0.0$ & $0.92 \pm 0.0$ & $0.88 \pm 0.1$ & $0.00 \pm 0.0$ & $0.00 \pm 0.0$ & $0.00 \pm 0.0$ & $11694 \pm 1439$ & $30369 \pm 4828$ & -- \\
        Choose Starter & $3.43 \pm 1.7$ & $0.79 \pm 0.4$ & $0.79 \pm 0.4$ & $0.75 \pm 0.4$ & $0.29 \pm 0.4$ & $0.00 \pm 0.0$ & $0.00 \pm 0.0$ & $6485 \pm 578$ & $20279 \pm 579$ & -- \\
        Choose Starter${~-~R_{heal}}$ & $3.40 \pm 0.4$ & $0.97 \pm 0.0$ & $0.96 \pm 0.0$ & $0.45 \pm 0.3$ & $0.01 \pm 0.0$ & $0.00 \pm 0.0$ & $0.00 \pm 0.0$ & $13871 \pm 2991$ & $33565 \pm 13303$ & -- \\
        Bulbasaur${~-~R_{heal}}$ & $3.18 \pm 0.3$ & $0.97 \pm 0.0$ & $0.95 \pm 0.0$ & $0.26 \pm 0.2$ & $0.00 \pm 0.0$ & $0.00 \pm 0.0$ & $0.00 \pm 0.0$ & $13433 \pm 950$ & $45943 \pm 3569$ & -- \\
        Charmander & $3.03 \pm 1.6$ & $0.69 \pm 0.4$ & $0.67 \pm 0.3$ & $0.53 \pm 0.4$ & $0.02 \pm 0.0$ & $0.17 \pm 0.3$ & $0.02 \pm 0.0$ & $7029 \pm 1132$ & $21309 \pm 5915$ & $48544 \pm 0$ \\
        Bulbasaur & $2.89 \pm 0.1$ & $0.95 \pm 0.0$ & $0.94 \pm 0.0$ & $0.00 \pm 0.0$ & $0.00 \pm 0.0$ & $0.00 \pm 0.0$ & $0.00 \pm 0.0$ & $9929 \pm 841$ & -- & -- \\
        Baseline$~+~10R_{nav}$ & $1.04 \pm 0.1$ & $0.04 \pm 0.1$ & $0.00 \pm 0.0$ & $0.00 \pm 0.0$ & $0.00 \pm 0.0$ & $0.00 \pm 0.0$ & $0.00 \pm 0.0$ & $11559 \pm 0$ & -- & -- \\
        Baseline${~-~R_{nav}}$ & $0.00 \pm 0.0$ & $0.00 \pm 0.0$ & $0.00 \pm 0.0$ & $0.00 \pm 0.0$ & $0.00 \pm 0.0$ & $0.00 \pm 0.0$ & $0.00 \pm 0.0$ & -- & -- & -- \\
        \bottomrule
        \end{tabular}

        }
        
        \label{tab:performance}
    \end{table}

\vfill  

\begin{table}[p]
        \centering
        \caption{Policy metrics describing the agent's behavior. Rows are ordered as in Table \ref{tab:performance}. The best value in each column is highlighted in bold. The species entropy refers to the probability distribution of catching distinct species. As the Fast agent usually does not catch Pokémon, but may evolve Squirtle to Wartortle, the entropy is lowest for these runs. The human playthroughs are more diverse in their party setup as indicated by the largest entropy.}

        \resizebox{\linewidth}{!}{  
        
\begin{tabular}{lllllllll}
        \toprule
        Experiment & Events Completed & Poké Center Visits & Num. Heals & Black Outs & Max Party Level & Visited Coordinates & Party Size & Species Entropy \\
        \midrule
        Human & $50.47 \pm 9.1$ & $12.53 \pm 7.7$ & $22.97 \pm 11.3$ & $2.23 \pm 2.1$ & $26.20 \pm 2.1$ & $1997 \pm 568$ & $2.07 \pm 1.2$ & $3.53 \pm 0.0$ \\ \hline
        Baseline${~-~R_{lvl}}$ & $\mathbf{57.73 \pm 11.4}$ & $20.29 \pm 24.2$ & $11.73 \pm 4.0$ & $9.01 \pm 3.2$ & $27.45 \pm 2.8$ & $4560 \pm 1015$ & $1.58 \pm 0.5$ & $1.24 \pm 0.2$ \\
        Fast & $57.21 \pm 5.4$ & $\mathbf{111.46 \pm 133.9}$ & $47.98 \pm 66.6$ & $9.68 \pm 1.7$ & $\mathbf{29.43 \pm 1.9}$ & $3392 \pm 744$ & $3.49 \pm 0.2$ & $2.54 \pm 0.1$ \\
        Baseline & $51.78 \pm 2.1$ & $19.09 \pm 19.6$ & $12.81 \pm 5.9$ & $9.08 \pm 2.5$ & $27.52 \pm 3.2$ & $4275 \pm 1406$ & $3.48 \pm 0.3$ & $\mathbf{2.61 \pm 0.1}$ \\
        Baseline${~-~R_{heal}}$ & $43.21 \pm 2.0$ & $0.04 \pm 0.1$ & $1.45 \pm 0.7$ & $9.74 \pm 1.2$ & $23.38 \pm 1.1$ & $\mathbf{5637 \pm 368}$ & $\mathbf{3.72 \pm 0.2}$ & $2.27 \pm 0.2$ \\
        GRU & $47.20 \pm 21.3$ & $18.16 \pm 25.3$ & $23.09 \pm 23.3$ & $8.25 \pm 5.2$ & $22.90 \pm 9.4$ & $2534 \pm 1421$ & $2.72 \pm 0.9$ & $1.87 \pm 1.0$ \\
        Fast${-R_{heal}}$ & $41.79 \pm 5.3$ & $0.01 \pm 0.0$ & $2.70 \pm 1.1$ & $9.27 \pm 1.8$ & $23.73 \pm 0.8$ & $5135 \pm 843$ & $3.20 \pm 0.3$ & $2.16 \pm 0.2$ \\
        Charmander${~-~R_{heal}}$ & $45.29 \pm 7.2$ & $0.01 \pm 0.0$ & $2.68 \pm 1.9$ & $8.54 \pm 1.6$ & $24.61 \pm 2.5$ & $5399 \pm 511$ & $3.32 \pm 0.3$ & $2.09 \pm 0.2$ \\
        Choose Starter & $48.42 \pm 17.7$ & $44.32 \pm 36.1$ & $33.75 \pm 34.1$ & $9.14 \pm 5.7$ & $24.72 \pm 8.7$ & $2577 \pm 1160$ & $1.79 \pm 0.4$ & $1.25 \pm 0.6$ \\
        Choose Starter${~-~R_{heal}}$ & $41.49 \pm 3.0$ & $0.01 \pm 0.0$ & $5.75 \pm 2.4$ & $9.73 \pm 2.2$ & $24.88 \pm 2.2$ & $4527 \pm 371$ & $1.86 \pm 0.8$ & $1.30 \pm 0.6$ \\
        Bulbasaur${~-~R_{heal}}$ & $36.55 \pm 2.6$ & $0.01 \pm 0.0$ & $8.74 \pm 4.6$ & $10.00 \pm 0.7$ & $21.80 \pm 0.8$ & $4787 \pm 406$ & $3.43 \pm 0.6$ & $2.34 \pm 0.2$ \\
        Charmander & $46.52 \pm 19.7$ & $57.83 \pm 57.8$ & $25.87 \pm 19.0$ & $6.01 \pm 3.5$ & $22.71 \pm 8.6$ & $2871 \pm 1415$ & $2.95 \pm 1.0$ & $2.22 \pm 1.1$ \\
        Bulbasaur & $30.32 \pm 2.2$ & $88.15 \pm 20.7$ & $\mathbf{399.33 \pm 177.1}$ & $7.76 \pm 3.5$ & $20.63 \pm 3.8$ & $2283 \pm 279$ & $3.17 \pm 0.5$ & $2.37 \pm 0.3$ \\
        Baseline$~+~10R_{nav}$ & $16.39 \pm 1.2$ & $0.00 \pm 0.0$ & $0.02 \pm 0.0$ & $3.49 \pm 2.5$ & $9.05 \pm 1.7$ & $3284 \pm 364$ & $1.03 \pm 0.0$ & $0.18 \pm 0.2$ \\
        Baseline${~-~R_{nav}}$ & $13.40 \pm 1.2$ & $0.00 \pm 0.0$ & $0.00 \pm 0.0$ & $\mathbf{0.00 \pm 0.0}$ & $6.00 \pm 0.0$ & $35 \pm 15$ & $1.00 \pm 0.0$ & $0.00 \pm 0.0$ \\
        \bottomrule
        \end{tabular}

        }

        \label{tab:policy}
    \end{table}

\end{landscape}

\clearpage
\noindent
an additional reward signal to incentivize all the behaviors associated with Cut. Indeed, many such challenges exist in Pokémon Red, and further refining the reward function risks introducing other vulnerabilities to be exploited and side effects to be realized, highlighting the environment's complexity. 

\subsubsection{Limitations}
\label{sec:reward_shaping}

While Pokémon Red serves as a valuable research environment, it comes with several limitations.  
First, the game is closed-source and requires owning a legal copy.  
Once a digital copy is available for training, additional engineering effort is needed to extract relevant information from the Game Boy emulator's RAM.

Second, the prolonged episode length significantly increases evaluation time.  
Running evaluation episodes takes as much wall-time as the training runs themselves, with a single training run lasting approximately 36 hours.  
This extended horizon also presents challenges for agents utilizing Recurrent Neural Networks.  
Processing long observation sequences—2048 steps in our case—leads to substantial VRAM overhead.
As such, in lieu of additional optimization, the GRU experiments were run on CPU only, with an average runtime of 24 days per training run.  

Furthermore, the vast number of possible gameplay strategies contributes to the curse of dimensionality.  
Comparing agents across numerous factors makes comprehensive evaluation labor-intensive.  
For example, analyzing item usage and move selection falls outside the scope of this paper, as does manually reviewing thousands of episodes.

Addressing these limitations may require refining the toolset, which has been effective for other environments.

\section{Related Work}

To our knowledge, no peer-reviewed publications have explored training agents on Pokémon Red or its successors.  
Recent works have, however, introduced Large Language Model (LLM) agents.
A preprint presented an agent focused solely on Pokémon battles, achieving human-parity performance \cite{hu2024pokellmonhumanparityagentpokemon}.
The startup nunu.ai demonstrated an agent that reached the third badge in Pokémon Emerald, relying on human intervention and a pathfinding algorithm.
More recently, Anthropic announced efforts to develop an agent capable of earning the third badge in Pokémon Red \cite{Anthropic2025}.
This is particularly noteworthy, as Gym 3 requires both using Cut and solving a hidden button puzzle.
Although details on these agents remain limited, the increasing interest in Pokémon as an LLM testbed highlights its growing relevance in AI research.
A promising direction for future work is an agent similar to Minecraft Voyager \cite{wang2023voyager}, which generates and evaluates code to perform actions.




\section{Conclusion and Outlook}

Our simple DRL baseline trains agents to complete an initial portion of Pokémon Red, serving as stepping stone for tackling its extensive challenges.
Key tasks like cutting trees and solving puzzles remain unexplored by our agents but pose significant hurdles, potentially requiring engineered solutions such as reward shaping or curriculum learning, thus further diverging from human perception and play.
We did not address generalization, as agents started from a fixed state while influencing RNG through their actions.
ROM randomizers could improve robustness by varying start conditions, such as the choice of starter Pokémon.
For handling long horizons, hierarchical DRL may reduce unnecessary actions, while advanced exploration methods could replace naive navigation rewards and binary image observations.
Overall, we anticipate Pokémon Red will play an essential role in future research.

\section*{Acknowledgment}

This research is supported by computing time from the Paderborn Center for Parallel Computing and the Linux-HPC-Cluster at TU Dortmund.
We also thank Keelan Donovan and Sky (Discord username) for their discussions and reviews, PufferLib \cite{pufferlib2024} for RL expertise, and Ryan Sullivan and Joseph Suarez for manuscript review.

\bibliographystyle{IEEEtran}
\bibliography{IEEEabrv, bibliography.bib}

\begin{thebibliography}{10}
\providecommand{\url}[1]{#1}
\csname url@samestyle\endcsname
\providecommand{\newblock}{\relax}
\providecommand{\bibinfo}[2]{#2}
\providecommand{\BIBentrySTDinterwordspacing}{\spaceskip=0pt\relax}
\providecommand{\BIBentryALTinterwordstretchfactor}{4}
\providecommand{\BIBentryALTinterwordspacing}{\spaceskip=\fontdimen2\font plus
\BIBentryALTinterwordstretchfactor\fontdimen3\font minus \fontdimen4\font\relax}
\providecommand{\BIBforeignlanguage}[2]{{%
\expandafter\ifx\csname l@#1\endcsname\relax
\typeout{** WARNING: IEEEtran.bst: No hyphenation pattern has been}%
\typeout{** loaded for the language `#1'. Using the pattern for}%
\typeout{** the default language instead.}%
\else
\language=\csname l@#1\endcsname
\fi
#2}}
\providecommand{\BIBdecl}{\relax}
\BIBdecl

\bibitem{Berner2019Dota2}
C.~Berner, G.~Brockman, B.~Chan, V.~Cheung, P.~Debiak, C.~Dennison, D.~Farhi, Q.~Fischer, S.~Hashme, C.~Hesse, R.~J{\'{o}}zefowicz, S.~Gray, C.~Olsson, J.~Pachocki, M.~Petrov, H.~P. de~Oliveira~Pinto, J.~Raiman, T.~Salimans, J.~Schlatter, J.~Schneider, S.~Sidor, I.~Sutskever, J.~Tang, F.~Wolski, and S.~Zhang, ``Dota 2 with large scale deep reinforcement learning,'' \emph{CoRR}, vol. abs/1912.06680, 2019.

\bibitem{kuettler2020nethack}
H.~K{\"{u}}ttler, N.~Nardelli, A.~H. Miller, R.~Raileanu, M.~Selvatici, E.~Grefenstette, and T.~Rockt{\"{a}}schel, ``{The NetHack Learning Environment},'' in \emph{NIPS}, 2020.

\bibitem{minedojo2022}
L.~Fan, G.~Wang, Y.~Jiang, A.~Mandlekar, Y.~Yang, H.~Zhu, A.~Tang, D.~Huang, Y.~Zhu, and A.~Anandkumar, ``Minedojo: Building open-ended embodied agents with internet-scale knowledge,'' in \emph{NIPS}, 2022.

\bibitem{pdubsYoutube}
\BIBentryALTinterwordspacing
P.~Whidden, ``Training ai to play pokemon with reinforcement learning,'' 2023, accessed: 2025-02-25. [Online]. Available: \url{https://www.youtube.com/watch?v=DcYLT37ImBY}
\BIBentrySTDinterwordspacing

\bibitem{Schulman2017}
J.~Schulman, F.~Wolski, P.~Dhariwal, A.~Radford, and O.~Klimov, ``Proximal policy optimization algorithms,'' \emph{CoRR}, vol. abs/1707.06347, 2017.

\bibitem{sutton2018reinforcement}
R.~S. Sutton and A.~G. Barto, \emph{Reinforcement Learning: An Introduction}, 2nd~ed.\hskip 1em plus 0.5em minus 0.4em\relax MIT Press, 2018.

\bibitem{pleines2023memory}
M.~Pleines, M.~Pallasch, F.~Zimmer, and M.~Preuss, ``Memory gym: Partially observable challenges to memory-based agents,'' in \emph{ICLR}, 2023.

\bibitem{mnih2015DQN}
V.~Mnih, K.~Kavukcuoglu, D.~Silver, A.~A. Rusu, J.~Veness, M.~G. Bellemare, A.~Graves, M.~A. Riedmiller, A.~Fidjeland, G.~Ostrovski, S.~Petersen, C.~Beattie, A.~Sadik, I.~Antonoglou, H.~King, D.~Kumaran, D.~Wierstra, S.~Legg, and D.~Hassabis, ``Human-level control through deep reinforcement learning,'' \emph{Nature}, vol. 518, pp. 529--533, 2015.

\bibitem{Pathak2017}
D.~Pathak, P.~Agrawal, A.~A. Efros, and T.~Darrell, ``Curiosity-driven exploration by self-supervised prediction,'' in \emph{ICML}, 2017.

\bibitem{Burda2019}
Y.~Burda, H.~Edwards, A.~J. Storkey, and O.~Klimov, ``Exploration by random network distillation,'' in \emph{ICLR}, 2019.

\bibitem{zhou2024}
H.~Zhou, A.~Huang, K.~Azizzadenesheli, D.~Childers, and Z.~Lipton, ``Timing as an action: Learning when to observe and act,'' in \emph{AISTATS}, 2024.

\bibitem{hafner2022deep}
D.~Hafner, K.-H. Lee, I.~Fischer, and P.~Abbeel, ``Deep hierarchical planning from pixels,'' in \emph{NIPS}, 2022.

\bibitem{hu2024pokellmonhumanparityagentpokemon}
S.~Hu, T.~Huang, and L.~Liu, ``Pokellmon: A human-parity agent for pokemon battles with large language models,'' \emph{CoRR}, vol. abs/2402.01118, 2024.

\bibitem{Anthropic2025}
\BIBentryALTinterwordspacing
Anthropic, ``Claude’s extended thinking,'' 2025, accessed: 2025-02-25. [Online]. Available: \url{https://www.anthropic.com/news/visible-extended-thinking}
\BIBentrySTDinterwordspacing

\bibitem{wang2023voyager}
G.~Wang, Y.~Xie, Y.~Jiang, A.~Mandlekar, C.~Xiao, Y.~Zhu, L.~Fan, and A.~Anandkumar, ``Voyager: An open-ended embodied agent with large language models,'' \emph{CoRR}, vol. abs/2305.16291, 2023.

\bibitem{pufferlib2024}
\BIBentryALTinterwordspacing
J.~Suarez, ``Pufferlib: Making reinforcement learning libraries and environments play nice,'' 2024, accessed on \today. [Online]. Available: \url{https://arxiv.org/abs/2406.12905}
\BIBentrySTDinterwordspacing

\end{thebibliography}

\end{document}